\documentclass[10pt,twocolumn,letterpaper]{article}

\usepackage{ijcb}
\usepackage{times}
\usepackage{epsfig}
\usepackage{graphicx, scalerel}
\usepackage{amsmath}
\usepackage{amssymb}

\usepackage{booktabs}
\usepackage[hyphens]{url}
\usepackage{hyperref}
\usepackage[hyphenbreaks]{breakurl}
\usepackage{multirow}
\usepackage{array}
\usepackage{tabu}
\usepackage{xcolor}

\newcommand\sbullet[1][.5]{\mathbin{\ThisStyle{\vcenter{\hbox{%
  \scalebox{#1}{$\SavedStyle\bullet$}}}}}%
}

\ijcbfinalcopy 

\ifijcbfinal\fi
\begin{document}

\title{Fingerprint Spoof Detection: Temporal Analysis of Image Sequence}

\author{Tarang Chugh and Anil K. Jain\\
Department of Computer Science and Engineering\\
Michigan State University, East Lansing, Michigan 48824\\
{\tt\small \{chughtar, jain\}@cse.msu.edu}}

\maketitle
\thispagestyle{empty}

\begin{abstract}
We utilize the dynamics involved in the imaging of a fingerprint on a touch-based fingerprint reader, such as perspiration, changes in skin color (blanching), and skin distortion, to differentiate real fingers from spoof (fake) fingers. Specifically, we utilize a deep learning-based architecture (CNN-LSTM) trained end-to-end using sequences of minutiae-centered local patches extracted from ten color frames captured on a COTS fingerprint reader. A time-distributed CNN (MobileNet-v1) extracts spatial features from each local patch, while a bi-directional LSTM layer learns the temporal relationship between the patches in the sequence. Experimental results on a database of $26,650$ live frames from $685$ subjects ($1,333$ unique fingers), and $32,910$ spoof frames of $7$ spoof materials (with 14 variants) shows the superiority of the proposed approach in both known-material and cross-material (generalization) scenarios. For instance, the proposed approach improves the state-of-the-art cross-material performance from TDR of $81.65\%$ to $86.20\%$ @ FDR = $0.2\%$.
\end{abstract}


\section{Introduction}
Fingerprint recognition technology is now widely adopted across the globe for a plethora of applications, including international border crossing\footnote{\url{https://www.dhs.gov/obim}}, forensics\footnote{\url{https://www.fbi.gov/services/cjis}}, unlocking smartphones\footnote{\url{https://support.apple.com/en-us/HT201371}}, and national ID\footnote{\url{https://uidai.gov.in/}} programs. However, one of the most critical premise for this wide acceptance is that users have trust in the security of a fingerprint recognition system, namely protection of enrolled fingerprints (templates) and detection of fingerprint spoofs~\cite{marcel2019handbook}. In this paper, we focus on fingerprint spoof detection.

\begin{figure}[t]
\centering
\includegraphics[width=\linewidth]{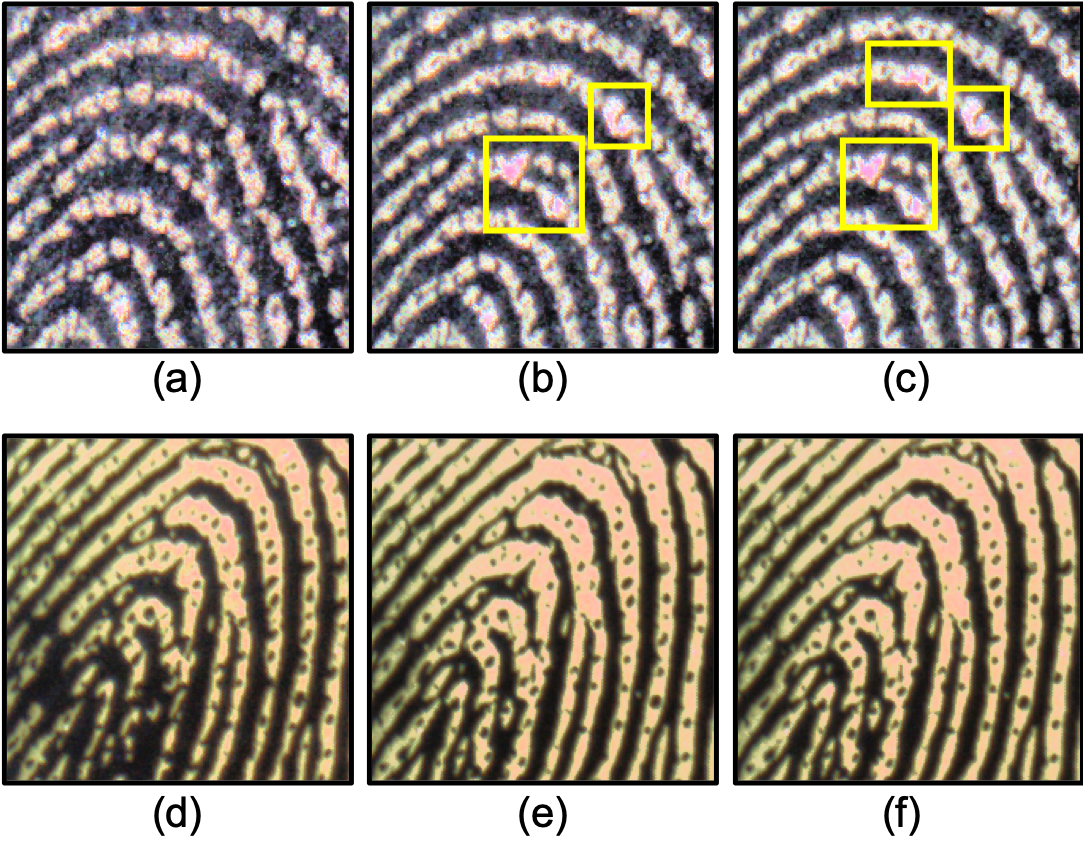}
\caption{A sequence of ten color frames are captured by a SilkID SLK20R fingerprint reader in quick succession ($8$ fps). The first, fifth, and tenth frames from a live (a) - (c), and spoof (tan pigmented third degree) (d) - (f) finger are shown here. Unlike spoofs, in the case of live fingers, appearance of sweat near pores (highlighted in yellow boxes) and changes in skin color (pinkish red to pale yellow) along the frames can be observed.}
\label{fig:framechanges}
\end{figure}

\begin{table*}[!htbp]
\caption{Studies primarily focused on fingerprint spoof detection using temporal analysis.}
\label{tab:litrev}
\centering
\resizebox{\linewidth}{!}{
\begin{tabular}{  p{3.5 cm}  >{\centering\arraybackslash}p{6.2 cm}  >{\centering\arraybackslash}p{4.9cm}  >{\centering\arraybackslash}p{5.3cm}  }
\toprule
\textbf{Study} & \textbf{Approach} & \textbf{Database} & \textbf{Performance} \\ \bottomrule

Parthasaradhi et al.~\cite{parthasaradhi2005time} &
Temporal analysis of perspiration pattern along friction ridges &
$1,840$ live from $33$ subjects and $1800$ spoof from $2$ materials, and $700$ cadaver from $14$ fingers &
Avg. Classification Accuracy = $90\%$ \\ \midrule

Kolberg et al.~\cite{kolberg2019multi} &
Blood flow detection using a sequence of $40$ Laser Speckle Contrast Images &
$1,635$ live from $163$ subjects and $675$ spoof images of 8 spoof materials (32 variants) &
TDR = $90.99\%$ @ FDR = $0.05\%$ \\ \midrule

Plesh et al.~\cite{pleshfingerprint} &
Fusion of static (LBP and CNN) and dynamic (changes in color ratio) features using a sequence of $2$ color frames &
$14,892$ live and $21,700$ spoof images of 10 materials &
TDR = $96.45\%$ (known-material) @ FDR = $0.2\%$ \\ \midrule

\textbf{Proposed Approach} &
Temporal analysis of minutiae-based local patch sequences from $10$ color frames using CNN + LSTM model &
$26,650$ live from $685$ subjects and $32,910$ spoof images of $7$ materials (14 variants) &
TDR = $\textbf{99.15\%}$ (known-material) and TDR = $\textbf{86.20\%}$ (cross-material) @ FDR = $0.2\%$ \\ \bottomrule

\end{tabular}
}
\end{table*}

Fingerprint spoof attacks\footnote{Fingerprint spoofs are one of the most common forms of presentation attacks (PA). The ISO standard IEC 30107-1:2016(E) defines presentation attacks as the \textit{``presentation to the biometric data capture subsystem with the goal of
interfering with the operation of the biometric system"}. Other forms of PAs include use of altered fingers and cadavers.} refer to finger-like artifacts with an accurate imitation of one's fingerprint fabricated for the purpose of stealing their identity. Techniques ranging from simple molding and casting to sophisticated 3D printing have been utilized to create spoofs with high fidelity~\cite{matsumoto2002impact, cao2016hacking, engelsma2018universal}. Various readily available and inexpensive materials (e.g. gelatin, wood glue, play doh) can be used to create spoofs that are capable of circumventing a fingerprint recognition system. For instance, in March 2018, a gang in Rajasthan (India) bypassed the biometric attendance systems, using wood glue spoofs casted in wax molds, to provide proxies for police academy entrance exams\footnote{\burl{https://www.medianama.com/2018/03/223-cloned-thumb-prints-used-to-spoof-biometrics-and-allow-proxies-to-answer-online-rajasthan-police-exam/}}. 

With the goal to detect such spoof attacks, various hardware and software-based spoof detection approaches have been proposed in literature~\cite{marcel2019handbook}. The hardware-based approaches typically utilize specialized sensors to detect the signs of vitality (blood flow, heartbeat, etc.) and/or sensing technologies for sub-dermal imaging~\cite{kolberg2019multi, tolosana2019biometric, chugh2019oct, moolla2019optical}. On the other hand, software-based approaches extract salient cues, related to anatomical (pores)~\cite{schuckers2017fingerprint} and texture-based features~\cite{xia2018novel}, from the captured fingerprint image(s). Chugh et al.~\cite{chugh2018fingerprint} utilized minutiae-based local patches to train deep neural networks that achieves state-of-the-art spoof detection performance. Gonzalez-Soler et al.~\cite{gonzalez2019fingerprint} proposed fusion of feature encodings of dense-SIFT features for robust spoof detection. 

Software-based approaches can be further classified into static and dynamic approaches based on the input. A static approach extracts discriminative spatial features from a single fingerprint image, while a dynamic approach utilizes an image sequence to extract spatial and/or temporal features for spoof detection.  For a comprehensive review on the existing static approaches, readers are referred to~\cite{marcel2019handbook, ghiani2017review}. 

In the case of dynamic approaches, published studies utilize temporal analysis to capture the physiological features, such as perspiration~\cite{parthasaradhi2005time, marasco2012combining}, blood flow~\cite{yau2007fake, kolberg2019multi}, skin distortion~\cite{antonelli2006fake}, and color change~\cite{yau2007fake, pleshfingerprint}. Table~\ref{tab:litrev} summarizes the dynamic approaches for fingerprint spoof detection reported in the literature. Some of the limitations of these studies include long capture time (2-5 seconds), expensive hardware, and/or small number of frames in the sequence. Moreover, it is likely that some live fingers may not exhibit any of these dynamic phenomenons to separate them from spoofs. For instance, some dry fingers may not exhibit signs of perspiration during the finger presentation or a spoof may produce similar distortion characteristics as that of some live fingers. 

We posit that automatic learning, as opposed to hand-engineering, of the dynamic features involved in the presentation of a finger can provide more robust and highly discriminating cues to distinguish live from spoofs. In this study, we propose to use a CNN-LSTM architecture to learn the spatio-temporal features across different frames in a sequence. We utilize a sequence of minutiae-centered local patches extracted from ten colored frames captured by a COTS fingerprint reader, SilkID SLK20R\footnote{\url{https://www.zkteco.com/en/product\_detail/SLK20R.html}}, at $8$ fps to train the network in an end-to-end manner. The use of minutiae-based local patches has been shown to achieve state-of-the-art spoof detection performance compared to randomly selected local patches in static images~\cite{chugh2017fingerprint}. Additionally, using minutiae-based local patches provides a large amount of training data, $71,530$ minutiae-based patch sequences, compared to $5,956$ whole-frame sequences. 

{\flushleft The main contributions of this study are:}

\begin{itemize}
\item Utilized sequences of minutiae-based local patches to train a CNN-LSTM architecture with the goal of learning discriminative spatio-temporal features for fingerprint spoof detection. The local patches are extracted from a sequence of ten colored frames captured in quick succession ($8$ fps) using a COTS fingerprint reader, SilkID SLK20R.

\item Experimental results on a dataset of $26,650$ live captures from $685$ subjects ($1333$ unique fingers) and $32,930$ spoof frames from 7 spoof materials (with 14 variants) shows that the proposed approach is able to improve the state-of-the-art cross-material performance from TDR of $81.65\%$ to $86.20\%$ @ FDR = $0.2\%$.

\end{itemize}

\begin{figure*}[htbp]
\centering
\includegraphics[width=0.89\linewidth]{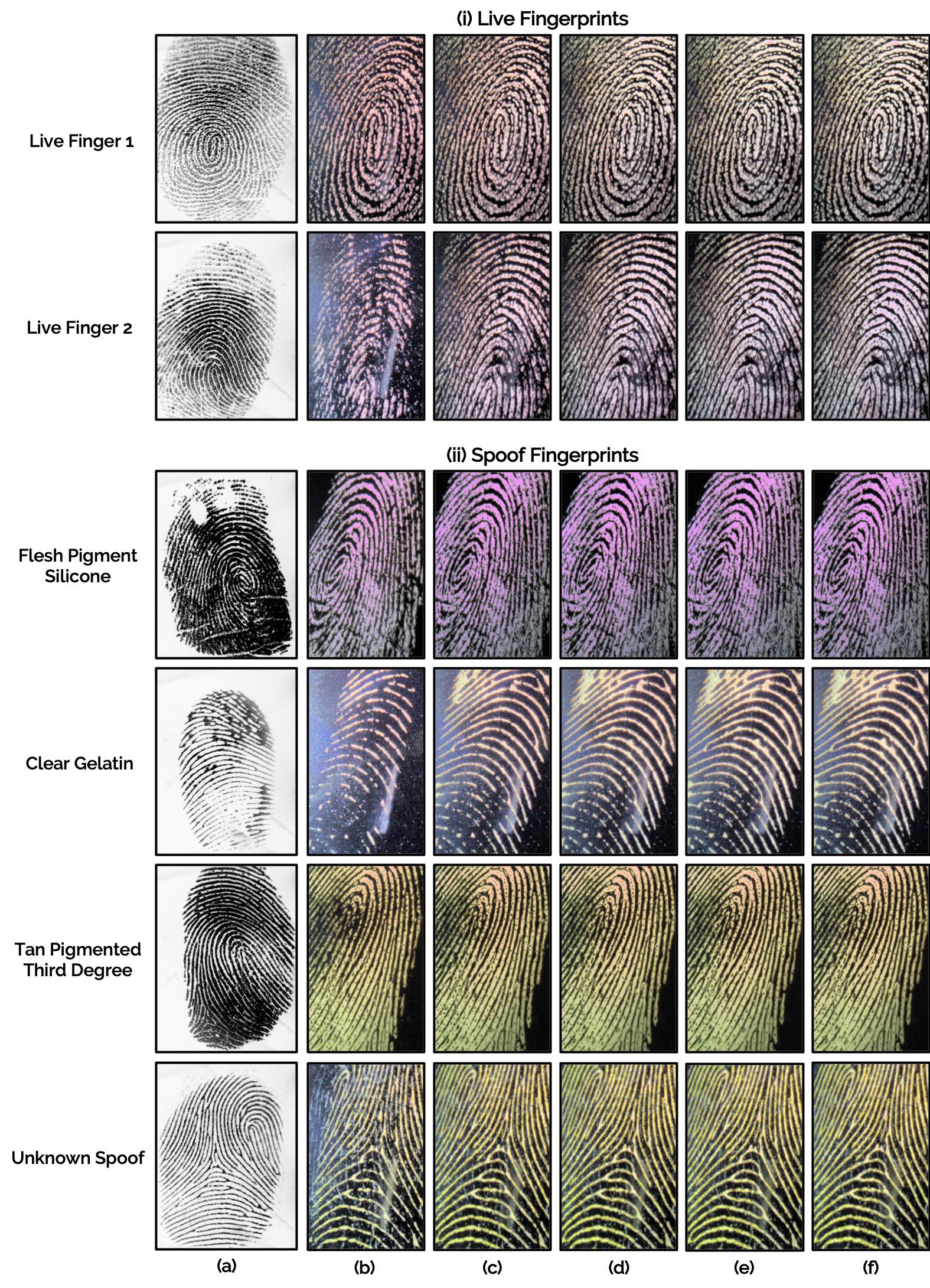}
\caption{Examples of (i) live and (ii) spoof fingerprint images. (a) Grayscale $1000$ ppi image, and (c)-(g) the first five (colored) frames captured by SilkID SLK20R Fast Frame Rate reader. Live frames exhibit the phenomenon of blanching of the skin, \textit{i.e.} displacement of blood when a live finger is pressed on the glass platen changing the finger color from red/pink to pale white.}
\label{fig:frame_samples}
\end{figure*}

\section{Proposed Approach}
\begin{figure*}[t]
\centering
\includegraphics[width=0.94\linewidth]{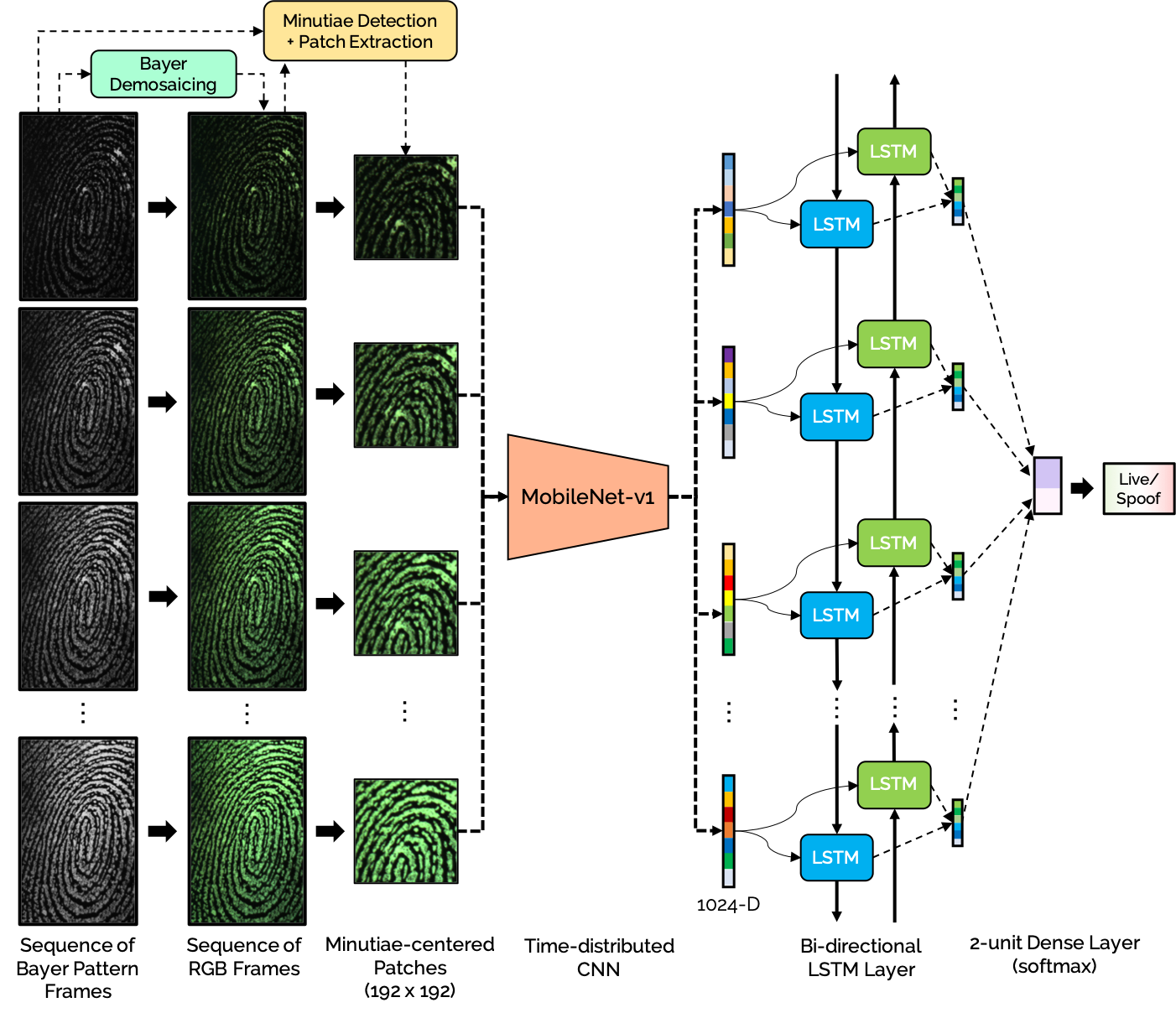}
\caption{An overview of the proposed approach utilizing a CNN-LSTM model trained end-to-end on sequences of minutiae-centered local patches for fingerprint spoof detection.}
\label{fig:overview}
\end{figure*}

The proposed approach consists of: (a) detecting minutiae from each of the frames and selecting the frame with the highest number of minutiae as the reference frame, (b) preprocessing the sequence of frames to convert them from Bayer pattern grayscale images to RGB images, (c) extracting local patches ($192 \times 192$) from all ten frames based on the location of detected minutiae in the reference frame, and (c) end-to-end training of a CNN-LSTM architecture using the sequences of minutiae-centered patches extracted from the ten frames. While a time-distributed CNN network (MobileNet-v1) with shared weights extracts deep features from the local patches, a bidirectional LSTM layer is utilized to learn the temporal relationship between the features extracted from the sequence. An overview of the proposed approach is presented in Figure~\ref{fig:overview}.
 
\subsection{Minutia Detection}
When a finger (or spoof) is presented to the SilkID SLK20R fingerprint reader, it captures a sequence of ten color frames, $F = \{f_1, f_2,...,f_{10}\}$, at 8 frames per second\footnote{It takes an average of 1.25 seconds to capture a sequence of ten frames.} (fps) and a resolution of $1000$ ppi. While the complete sensing region ($h \times w$) in a SilkID fingerprint reader is $800 \times 600$ pixels, each of the ten colored frames are captured from a smaller central region of $630 \times 390$ pixels to ensure the fast frame rate of $8$ fps. The starting and ending frames in the sequence may have little or no friction ridge details if the finger is not yet completely placed or quickly removed from the reader, respectively. Therefore, we extract minutiae information from all of the ten frames using the algorithm proposed by Cao et al.~\cite{cao2019end}. Since the minutiae detector proposed in~\cite{cao2019end} is optimized for $500$ ppi fingerprint images, all frames are resized before extracting the minutiae. The frame with the maximum number of detected minutiae is selected as the reference frame ($f^{ref}$) and the corresponding minutiae set as the reference minutiae set ($M^{ref}$). 

\begin{figure}[t]
\centering
\includegraphics[width=\linewidth]{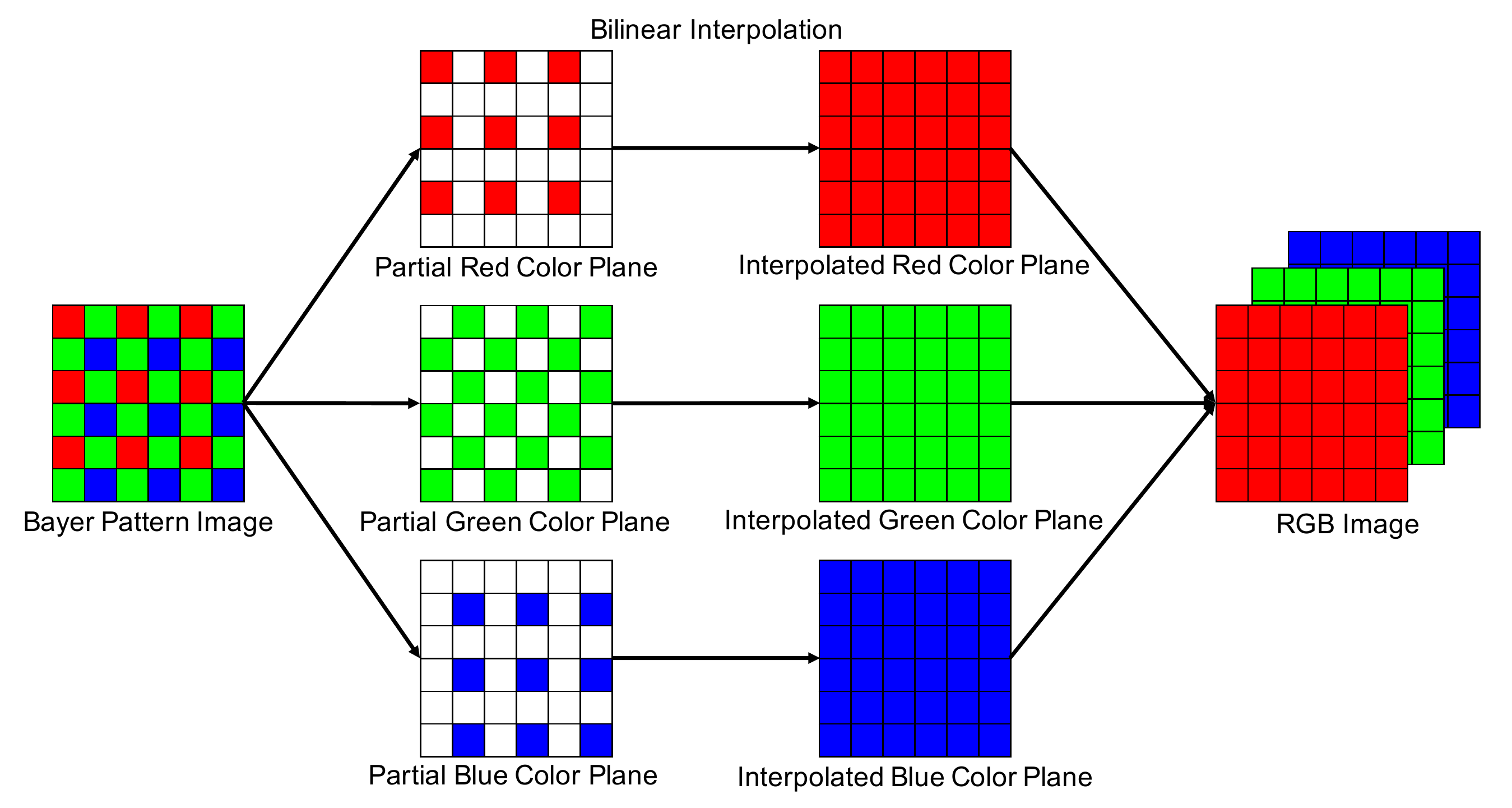}
\caption{A Bayer color filter array consists of alternating rows of red-green and green-blue filters. Bilinear interpolation of each channel is utilized to construct the RGB image.}
\label{fig:bayerfilter}
\vspace{-1mm}
\end{figure}

\subsection{Pre-processing}
A digital sensor, containing a large array of photo-sensitive sites (pixels), is typically used in conjunction with a color filter array to permit only particular colors of light at each pixel. SilkID fingerprint reader employs one of the most common filter arrays, called as \textit{Bayer filter array}, consisting of alternating rows of red-green (RG) and green-blue (GB) filters. \textit{Bayer demosaicing}~\cite{li2008image} (debayering) is the process of converting a bayer pattern image to an image with complete RGB color information at each pixel. It utilizes bilinear interpolation technique~\cite{thevenaz2000image} to estimate the missing pixels in the three color planes as shown in Figure~\ref{fig:bayerfilter}. The original sequence of grayscale Bayer pattern frames ($10 \times 630 \times 390$) is converted to RGB colorspace using an OpenCV~\cite{bradski2008learning} function, \textit{cv2.cvtColor()}, with the parameter $flag=$ \textit{cv2.COLOR\_BAYER\_BG2RGB}. After debayering, the frames have high pixel intensity values in the green channel (see Figure~\ref{fig:overview}) as SilkID readers are calibrated with strong gains on green pixels for generating high quality FTIR images. We utilize these raw images for our experiments. For visualization purposes, we reduce the green channel intensity values by a factor of $0.58$ and perform histogram equalization on intensity value in the HSV colorspace\footnote{Reducing gain in green channel and histogram equalization achieved similar or lower performance compared to using raw color images. Therefore, raw images were used for all experiments.} (see Figures~\ref{fig:framechanges} and~\ref{fig:frame_samples}).
\subsection{Local Patch Extraction}
For each of the detected minutiae from the reference frame, $m_i \in M^{ref}$, we extract a sequence of ten local patches, $P_i = \{p_i^{f_1}, p_i^{f_2},..., p_i^{f_{10}}\}$, of size $192 \times 192$, from the ten frames $(F)$, centered at the minutiae location\footnote{Minutiae coordinates extracted from the resized $500$ ppi frames are doubled to correspond to minutiae coordinates in the original $1000$ ppi frames.},~\textit{i.e.} $m_i = \{x_i, y_i\}$. This results in a total of $k$ patch sequences, where $k$ is equal to the number of detected minutiae in the reference frame. Chugh et al.~\cite{chugh2017fingerprint} reported that for $500$ ppi fingerprint images, the minutiae-based patches of size $96 \times 96$ pixels achieved the best performance compared to patch sizes of $64 \times 64$ pixels and $128 \times 128$ pixels. Therefore, for $1000$ ppi images in our case, we selected the patch size of $192 \times 192$ pixels to ensure similar amount of friction ridge area in each patch, as contained in $96 \times 96$ pixels patch size for $500$ ppi fingerprint images.. Each local patch from the reference frame is centered around the minutiae. However, this might not hold true for non-reference frames where the minutiae may shift due to non-linear distortion of human skin and non-rigid spoof materials. We hypothesize that the proposed approach can utilize the differences in the non-linear shift along the sequences of local patches as a salient cue to distinguish between live and spoof presentations.

\begin{table*}[htbp]
\centering
\caption{Summary of the fingerprint database utilized in this study.}
\label{tab:database}
\resizebox{\linewidth}{!}{
\begin{tabular}{ >{\raggedright\arraybackslash}p{7cm} >{\centering\arraybackslash}p{3.5cm} >{\raggedleft\arraybackslash}p{4.4cm}  >{\raggedleft\arraybackslash}p{4.4cm}} \toprule
\textbf{Spoof Material} & \textbf{Mold Type} & \textbf{\# Presentations} & \textbf{\# Frames} \\ \bottomrule

\multicolumn{2}{p{4cm}}{\textbf{Ecoflex silicone}}				&			& 						\\
\hspace{2mm}$\sbullet[.76]$ Ecoflex 00-35, flesh tone pigment 	& Dental		& $757$ 	& $7,570$ 		\\
\hspace{2mm}$\sbullet[.76]$ Ecoflex 00-50, flesh tone pigment 	& 3D Printed	& $138$ 	& $1,380$ 		\\
\hspace{2mm}$\sbullet[.76]$ Ecoflex 00-50, tan pigment		& 3D Printed 	& $130$	& $1,300$			\\
\midrule

\multicolumn{2}{p{4cm}}{\textbf{Gelatin}} & 			&									\\
\hspace{2mm}$\sbullet[.76]$ Ballistic gelatin, flesh tone dye & 3D Printed	& $50$ 	& $500$ 			\\
\hspace{2mm}$\sbullet[.76]$ Knox gelatin, clear		& 3D Printed	& $84$ 	& $840$ 			\\
\midrule

\textbf{Third degree silicone}						&			&		&				\\
\hspace{2mm}$\sbullet[.76]$ Light flesh tone pigment	& Dental		& $131$	& $1,310$			\\
\hspace{2mm}$\sbullet[.76]$ Tan pigment				& Dental		& $98$	& $980$			\\
\hspace{2mm}$\sbullet[.76]$ Beige suede powder		& Dental		& $43$	& $430$			\\
\hspace{2mm}$\sbullet[.76]$ Medium flesh tone pigment	& Dental		& $36$	& $360$			\\
\midrule

\textbf{Crayola Model Magic}						&			& 		&				\\
\hspace{2mm}$\sbullet[.76]$ White color				& Dental		& $910$	& $9,100$			\\
\hspace{2mm}$\sbullet[.76]$ Red color				& Dental		& $308$	& $3,080$			\\
\midrule

\textbf{Pigmented Dragon Skin (flesh tone)}			& Dental		& $452$ 	& $4,520$			\\ \midrule
\textbf{Conductive Silicone}						& 3D Printed	& $87$	& $870$			\\ \midrule
\raggedright\textbf{Unknown Spoof (JHU-APL)}			& 3D Printed	& $67$ 	& $670$			\\ \toprule

\textbf{Total Spoofs} 								& 	& $\textbf{3,291}$		& $\textbf{32,910}$ \\ \midrule
\multicolumn{2}{p{4cm}}{\textbf{Total Lives} ($685$ subjects)} 	& $\textbf{2,665}$		& $\textbf{26,650}$ \\ \bottomrule
\end{tabular}
}
\vspace{-1mm}
\end{table*}

\subsection{Network Architecture}
Several deep Convolutional Neural Network (CNN) architectures, such as VGG~\cite{simonyan2014very}, Inception-v3~\cite{szegedy2016rethinking}, MobileNet-v1~\cite{howard2017mobilenets} etc., have been shown to achieve state-of-the-art performance for many vision-based tasks, including fingerprint spoof detection~\cite{nogueira2016fingerprint, chugh2017fingerprint}. Unlike traditional approaches where spatial filters are hand-engineered, CNNs can automatically learn salient features from the given image databases. However, as CNNs are feed-forward networks, they are not well-suited to capture the temporal dynamics involved in a sequence of images. On the other hand, a Recurrent Neural Network (RNN) architecture with feedback connections can process a sequence of data to learn the temporal features.

With the goal of learning highly discriminative and generalizable spatio-temporal features for fingerprint spoof detection, we utilize a joint CNN-RNN architecture that can extract deep spatial features from each frame, and learn the temporal relationship across the sequence. One of the most popular RNN architectures is Long Short-Term Memory~\cite{hochreiter1997long} that can learn long range dependencies from the input sequences. The proposed network architecture utilizes a time-distributed MobileNet-v1 CNN architecture followed by a Bi-directional LSTM layer\footnote{Experiments with uni-directional LSTM layer achieved lower or similar performance compared to when using bi-directional layer.} and a 2-unit softmax layer for the binary classification problem \textit{i.e.} live vs. spoof. See Figure~\ref{fig:overview}.

MobileNet-v1 is a low-latency network with only $4.24$M trainable parameters compared to other networks, such as Inception-v3 ($23.2$M) and VGG ($138$M), which achieve comparable performance in large-scale vision tasks~\cite{russakovsky2015imagenet}. In low resource requirements such as smartphones and embedded devices, MobileNet-v1 is well-suited for real-time spoof detection. Most importantly, it has been shown to achieve state-of-the-art performance for fingerprint spoof detection~\cite{chugh2018fingerprint} on publicly available datasets~\cite{ghiani2017review}. It takes an input image of size $224 \times 224 \times 3$, and outputs a 1024-dimensional feature vector (bottleneck layer). We resize the local patches from $192 \times 192$ to $224 \times 224$ as required by the MobileNet-v1 input. For the purposes of processing a sequence of images, we utilize a Keras' TimeDistributed wrapper to utilize the MobileNet-v1 architecture as a feature extractor with shared parameters across different frames (time-steps) in the sequence.

\begin{table}[t]
\caption{Performance comparison (TDR (\%) @ FDR = 0.2\%) between the proposed approach and the state-of-the-art~\cite{chugh2018fingerprint} for known-material scenario, where the spoof materials used in testing are also known during training.}
\label{tab:knownresults}
\centering
\resizebox{\linewidth}{!}{
\begin{tabular}{ >{\arraybackslash}p{4.4 cm}  >{\centering\arraybackslash}p{2.2 cm}  >{\centering\arraybackslash}p{3.5 cm}}
\toprule
\textbf{Approach} & \textbf{Architecture} & \textbf{TDR (\%) ($\pm$ s.d.) @ FDR = 0.2\%} \\ 
 \bottomrule
 
\textbf{Still (Whole Image)} 							& CNN 		& 96.90 $\pm$ 0.78 		  \\
\textbf{Still (Minutiae Patches)~\cite{chugh2018fingerprint}} 	& CNN 		& 99.11 $\pm$ 0.24 		  \\
\textbf{Sequence (Whole Frames)} 						& CNN-LSTM	& 98.93 $\pm$ 	0.44		   \\
\textbf{Sequence (Minutiae Patches)}					& CNN-LSTM	& \textbf{99.25 $\pm$ 0.22}  \\

\bottomrule
\end{tabular}
}
\end{table}

\begin{table*}[t]
\caption{Performance comparison (TDR (\%) @ FDR = 0.2\%) between the proposed approach and the state-of-the-art~\cite{chugh2018fingerprint} for three cross-material scenarios, where the spoof materials used in testing are unknown during training.}
\label{tab:crossresults}
\centering
\resizebox{\linewidth}{!}{
\begin{tabular}{  p{3.1 cm}  >{\centering\arraybackslash}p{3.5 cm}  >{\centering\arraybackslash}p{3.5 cm}  >{\centering\arraybackslash}p{3.5 cm} >{\centering\arraybackslash}p{4.9 cm}}
\toprule
 & \multicolumn{2}{p{7cm}}{\centering\textbf{Baselines}} & \multicolumn{2}{p{7.9cm}}{\centering\textbf{Proposed Approach}} \\ \midrule
\textbf{Unknown Material} & \textbf{Whole Image (Grayscale)} & \textbf{Fingerprint Spoof Buster~\cite{chugh2018fingerprint}} & \textbf{Sequence of Whole Images} & \textbf{Sequence of Minutiae-based Patches} \\
 \bottomrule
\textbf{Third Degree} & 43.83 	& 79.20	& 80.44 	& 84.50 	\\
\textbf{Gelatin} 		& 50.74 	& 76.52 	& 73.88 	& 82.81	\\
\textbf{Ecoflex} 		& 77.37	& 89.23	& 87.55 	& 91.28 	\\
\bottomrule
\textbf{Mean $\pm$ s.d.} & 57.31 $\pm$ 17.71 & $81.65 \pm 6.70$ & $80.62 \pm 6.84$ & $\textbf{86.20 $\pm$ 4.48}$  \\
\bottomrule
\end{tabular}
}
\end{table*}


\subsection{Implementation Details}
The network architecture is designed in the Keras framework\footnote{\url{https://keras.io/}} and trained from scratch on a Nvidia GTX 1080Ti GPU. We utilize the MobileNet-v1 architecture without its last layer wrapped in a Time-Distributed layer. The Bi-directional LSTM layer contains 256 units and has a dropout rate of $0.25$. We utilize Adam~\cite{kingma2014adam} optimizer with a learning rate of $0.001$ and a binary cross entropy loss function. The network is trained end-to-end with a batch size of $4$. The network is trained for $80$ epochs with early-stopping and $patience$ = $20$.

\section{Experimental Results}
\subsection{Database}
\label{sec:database}
In this study, we utilize a large-scale fingerprint database of $26,650$ live frames from $685$ subjects, and $32,930$ spoof frames of $7$ materials (14 variants) collected on SilkID SLK20R fingerprint reader. This database is constructed by combining fingerprint images collected from two sources. First, as part of the IARPA ODIN program~\cite{IARPAProject}, a large-scale Government Controlled Test (GCT-3) was conducted at Johns Hopkins University Applied Physics Laboratory (JHUAPL) facility in Nov. 2019, where a total of $685$ subjects with diverse demographics (in terms of age, profession, gender, and race) were recruited to present their real (live) as well as spoof biometric data (fingerprint, face, and iris). The spoof fingerprints were fabricated using 5 different spoof materials (11 variants) and a variety of fabrication techniques, including use of dental and 3D printed molds. For a balanced live and spoof data distribution, we utilize only right thumb and right index fingerprint images for the live data. Second, we collected spoof data in a lab setting\footnote{This database will be made accessible to the interested researchers after signing a license agreement.} using dental molds casted with three different materials, namely ecoflex (with flesh tone pigment), crayola model magic (red and white colors), and dragon skin (with flesh tone pigment). The details of the combined database are summarized in Table~\ref{tab:database}. 

\subsection{Results}
To demonstrate the robustness of our proposed approach, we evaluate it under two different settings, namely \textit{Known-Material} and \textit{Cross-Material} scenarios.

\subsubsection{Known-Material Scenario}
In this scenario, the same set of spoof materials are included in the train and test sets. To evaluate this, we utilize five-fold cross validation splitting the live and spoof datasets in 80/20 splits for training and testing with no subject overlap. In each of the five folds, there are $21,320$ live and $26,400$ spoof frames in training and rest in testing. Table~\ref{tab:knownresults} presents the results achieved by the proposed approach on known-materials compared to a state-of-the-art approach~\cite{chugh2018fingerprint} that utilizes minutiae-based local patches from static grayscale images. The proposed approach improves the spoof detection performance from TDR of $99.11\%$ to $99.25\%$ @ FDR = 0.2\%.

\subsubsection{Cross-Material Scenario}
In this scenario, the spoof materials used in the test set are unknown during training. We simulate this scenario by adopting a leave-one-out protocol, where one material (including all its variants) is removed from training, and it used for evaluating the trained model. It is a more challenging and practical setting as it evaluates the generalizability of a spoof detector against spoofs that are never seen during training. For instance, in one of the cross-material experiments, we exclude all of the third degree spoofs (pigmented, tan, beige powder, and medium) from training, and use them for testing. The live data is randomly divided in a 80/20 split, with no subject overlap, for training and testing, respectively. The proposed approach improves the cross-material spoof detection performance from TDR of $81.65\%$ to $86.20\%$ @ FDR = $0.2\%$. Table~\ref{tab:crossresults} presents the spoof detection performance achieved by the proposed approach, on three cross-material experiments, compared to a state-of-the-art approach. 

\subsection{Processing Times}
The proposed network architecture takes around $4-6$ hours to converge when trained with sequences of whole frames, and $24-30$ hours with sequences of minutiae-based local patches, using a Nvidia GTX 1080Ti GPU. An average of 11 (13) sequences of minutiae-based local patches are extracted from the live (spoof) frames. The average classification time for a single presentation, including preprocessing, minutiae-detection, patch extraction, and sequence generation and inference, on a Nvidia GTX 1080 Ti GPU is $58$ms for full frame-based sequences, and $393$ms for minutiae-based patch sequences.

\section{Conclusions}
A robust and generalizable spoof detector is pivotal in the security and privacy of fingerprint recognition systems against unknown spoof attacks. In this study, we utilized a sequence of local patches centered at detected minutiae from ten color frames captured at $8$ fps as the finger is presented on the sensor. We posit that the dynamics involved in the presentation of a finger, such as skin blanching, distortion, and perspiration, provide discriminating cues to distinguish live from spoofs. We utilize a jointly learned CNN-LSTM model to learn the spatio-temporal dynamics across different frames in the sequence. The proposed approach improves the spoof detection performance from TDR of $99.11\%$ to $99.25\%$ @ FDR = $0.2\%$ in known-material scenarios, and from TDR of 81.65\% to 86.20\% @ FDR = 0.2\% in cross-material scenarios. In future, we will explore the use of live sequences to learn a one-class classifier for generalized fingerprint spoof detection.

\section{Acknowledgement}
This research is based upon work supported in part by the Office of the Director of National Intelligence (ODNI), Intelligence Advanced Research Projects Activity (IARPA), via IARPA R\&D Contract No. 2017 - 17020200004. The views and conclusions contained herein are those of the authors and should not be interpreted as necessarily representing the official policies, either expressed or implied, of ODNI, IARPA, or the U.S. Government. The U.S. Government is authorized to reproduce and distribute reprints for governmental purposes notwithstanding any copyright annotation therein.

{\small
\bibliographystyle{ieee}
\bibliography{egbib}

\begin{thebibliography}{10}\itemsep=-1pt

\bibitem{antonelli2006fake}
A.~Antonelli, R.~Cappelli, D.~Maio, and D.~Maltoni.
\newblock Fake finger detection by skin distortion analysis.
\newblock {\em IEEE Transactions on Information Forensics and Security},
  1(3):360--373, 2006.

\bibitem{bradski2008learning}
G.~Bradski and A.~Kaehler.
\newblock {\em Learning OpenCV: Computer vision with the OpenCV library}.
\newblock O'Reilly Media, Inc., 2008.

\bibitem{cao2016hacking}
K.~Cao and A.~K. Jain.
\newblock {Hacking mobile phones using 2D Printed Fingerprints}.
\newblock MSU Tech. report, MSU-CSE-16-2
  \url{https://www.youtube.com/watch?v=fZJI_BrMZXU}, 2016.

\bibitem{cao2019end}
K.~Cao, D.-L. Nguyen, C.~Tymoszek, and A.~K. Jain.
\newblock End-to-end latent fingerprint search.
\newblock {\em IEEE Transactions on Information Forensics and Security},
  15:880--894, 2019.

\bibitem{chugh2017fingerprint}
T.~Chugh, K.~Cao, and A.~K. Jain.
\newblock {Fingerprint Spoof Detection using Minutiae-based Local Patches}.
\newblock In {\em IEEE International Joint Conference on Biometrics (IJCB)},
  2017.

\bibitem{chugh2018fingerprint}
T.~Chugh, K.~Cao, and A.~K. Jain.
\newblock {Fingerprint Spoof Buster: Use of Minutiae-centered Patches}.
\newblock {\em IEEE Transactions on Information Forensics and Security},
  13(9):2190--2202, 2018.

\bibitem{chugh2019oct}
T.~Chugh and A.~K. Jain.
\newblock {OCT Fingerprints: Resilience to Presentation Attacks}.
\newblock {\em arXiv preprint arXiv:1908.00102}, 2019.

\bibitem{engelsma2018universal}
J.~J. Engelsma, S.~S. Arora, A.~K. Jain, and N.~G. Paulter.
\newblock {Universal 3D wearable Fingerprint Targets: Advancing Fingerprint
  Reader Evaluations}.
\newblock {\em IEEE Transactions on Information Forensics and Security},
  13(6):1564--1578, 2018.

\bibitem{ghiani2017review}
L.~Ghiani, D.~A. Yambay, V.~Mura, G.~L. Marcialis, F.~Roli, and S.~A.
  Schuckers.
\newblock {Review of the Fingerprint Liveness Detection (LivDet) competition
  series: 2009 to 2015}.
\newblock {\em Image and Vision Computing}, 58:110--128, 2017.

\bibitem{gonzalez2019fingerprint}
L.~J. Gonz{\'a}lez-Soler, M.~Gomez-Barrero, L.~Chang, A.~P{\'e}rez-Su{\'a}rez,
  and C.~Busch.
\newblock {Fingerprint Presentation Attack Detection Based on Local Features
  Encoding for Unknown Attacks}.
\newblock {\em arXiv preprint arXiv:1908.10163}, 2019.

\bibitem{hochreiter1997long}
S.~Hochreiter and J.~Schmidhuber.
\newblock Long short-term memory.
\newblock {\em Neural computation}, 9(8):1735--1780, 1997.

\bibitem{howard2017mobilenets}
A.~G. Howard, M.~Zhu, B.~Chen, D.~Kalenichenko, W.~Wang, T.~Weyand,
  M.~Andreetto, and H.~Adam.
\newblock {Mobilenets: Efficient Convolutional Neural Networks for Mobile
  Vision Applications}.
\newblock {\em arXiv preprint arXiv:1704.04861}, 2017.

\bibitem{kingma2014adam}
D.~P. Kingma and J.~Ba.
\newblock Adam: A method for stochastic optimization.
\newblock {\em arXiv preprint arXiv:1412.6980}, 2014.

\bibitem{kolberg2019multi}
J.~Kolberg, M.~Gomez-Barrero, and C.~Busch.
\newblock Multi-algorithm benchmark for fingerprint presentation attack
  detection with laser speckle contrast imaging.
\newblock In {\em IEEE International Conference of the Biometrics Special
  Interest Group (BIOSIG)}, pages 1--5, 2019.

\bibitem{li2008image}
X.~Li, B.~Gunturk, and L.~Zhang.
\newblock Image demosaicing: A systematic survey.
\newblock In {\em Visual Communications and Image Processing}, volume 6822.
  International Society for Optics and Photonics, 2008.

\bibitem{marasco2012combining}
E.~Marasco and C.~Sansone.
\newblock Combining perspiration-and morphology-based static features for
  fingerprint liveness detection.
\newblock {\em Pattern Recognition Letters}, 33(9), 2012.

\bibitem{marcel2019handbook}
S.~Marcel, M.~S. Nixon, J.~Fierrez, and N.~Evans, editors.
\newblock {\em "Handbook of Biometric Anti-Spoofing: Presentation Attack
  Detection"}.
\newblock Springer, 2 edition, 2019.

\bibitem{matsumoto2002impact}
T.~Matsumoto, H.~Matsumoto, K.~Yamada, and S.~Hoshino.
\newblock Impact of artificial gummy fingers on fingerprint systems.
\newblock In {\em Proc. SPIE}, volume 4677, pages 275--289, 2012.

\bibitem{moolla2019optical}
Y.~Moolla, L.~Darlow, A.~Sharma, A.~Singh, and J.~Van Der~Merwe.
\newblock Optical coherence tomography for fingerprint presentation attack
  detection.
\newblock In {\em Handbook of Biometric Anti-Spoofing}. Springer, 2019.

\bibitem{nogueira2016fingerprint}
R.~F. Nogueira, R.~de~Alencar~Lotufo, and R.~C. Machado.
\newblock {Fingerprint Liveness Detection Using Convolutional Neural Networks}.
\newblock {\em IEEE Transactions on Information Forensics and Security},
  11(6):1206--1213, 2016.

\bibitem{IARPAProject}
{ODNI, IARPA}.
\newblock {IARPA-BAA-16-04 (Thor)}.
\newblock
  \url{https://www.iarpa.gov/index.php/research-programs/odin/odin-baa}, 2016.

\bibitem{parthasaradhi2005time}
S.~T. Parthasaradhi, R.~Derakhshani, L.~A. Hornak, and S.~A. Schuckers.
\newblock {Time-series Detection of Perspiration as a Liveness Test in
  Fingerprint Devices}.
\newblock {\em IEEE Transactions on Systems, Man, and Cybernetics, Part C
  (Applications and Reviews)}, 35(3):335--343, 2005.

\bibitem{pleshfingerprint}
R.~Plesh, K.~Bahmani, G.~Jang, D.~Yambay, K.~Brownlee, T.~Swyka, P.~Biometrics,
  P.~Johnson, A.~Ross, and S.~Schuckers.
\newblock {Fingerprint Presentation Attack Detection utilizing Time-Series,
  Color Fingerprint Captures}.
\newblock In {\em IEEE International Conference on Biometrics (ICB)}, 2019.

\bibitem{russakovsky2015imagenet}
O.~Russakovsky, J.~Deng, H.~Su, J.~Krause, S.~Satheesh, S.~Ma, Z.~Huang,
  A.~Karpathy, A.~Khosla, M.~Bernstein, et~al.
\newblock Imagenet large scale visual recognition challenge.
\newblock {\em Proc. International Journal of Computer Vision (IJCV)},
  115(3):211--252, 2015.

\bibitem{schuckers2017fingerprint}
S.~Schuckers and P.~Johnson.
\newblock {Fingerprint Pore Analysis for Liveness Detection}, Nov.~14 2017.
\newblock {US Patent 9,818,020}.

\bibitem{simonyan2014very}
K.~Simonyan and A.~Zisserman.
\newblock Very deep convolutional networks for large-scale image recognition.
\newblock {\em arXiv preprint arXiv:1409.1556}, 2014.

\bibitem{szegedy2016rethinking}
C.~Szegedy, V.~Vanhoucke, S.~Ioffe, J.~Shlens, and Z.~Wojna.
\newblock {Rethinking the Inception Architecture for Computer Vision}.
\newblock In {\em Proc. IEEE CVPR}, pages 2818--2826, 2016.

\bibitem{thevenaz2000image}
P.~Th{\'e}venaz, T.~Blu, and M.~Unser.
\newblock Image interpolation and resampling.
\newblock {\em Handbook of medical imaging, processing and analysis},
  1(1):393--420, 2000.

\bibitem{tolosana2019biometric}
R.~Tolosana, M.~Gomez-Barrero, C.~Busch, and J.~Ortega-Garcia.
\newblock {Biometric Presentation Attack Detection: Beyond the Visible
  Spectrum}.
\newblock {\em IEEE Transactions on Information Forensics and Security}, 2019.

\bibitem{xia2018novel}
Z.~Xia, C.~Yuan, R.~Lv, X.~Sun, N.~N. Xiong, and Y.-Q. Shi.
\newblock {A Novel Weber Local Binary Descriptor for Fingerprint Liveness
  Detection}.
\newblock {\em IEEE Transactions on Systems, Man, and Cybernetics: Systems},
  2018.

\bibitem{yau2007fake}
W.-Y. Yau, H.-T. Tran, E.-K. Teoh, and J.-G. Wang.
\newblock Fake finger detection by finger color change analysis.
\newblock In {\em International Conference on Biometrics}, pages 888--896.
  Springer, 2007.

\end{thebibliography}
}

\end{document}